\documentclass[sigconf,nonacm]{acmart}

\usepackage{algorithm}
\usepackage{algorithmic}
\usepackage{multirow}
\usepackage{tikz}
\usetikzlibrary{
    shapes.geometric, 
    arrows.meta, 
    positioning, 
    calc, 
    backgrounds, 
    fit, 
    shadows.blur, 
    decorations.pathreplacing
}

\setcopyright{none}

\begin{document}

\title{DMamba: Decomposition-enhanced Mamba for Time Series Forecasting}


\author{Ruxuan Chen}
\authornote{Both authors contributed equally to this research.}
\email{rc1c24@soton.ac.uk}
\orcid{0009-0008-3858-5418}
\affiliation{   \institution{Harbin Engineering University}
  \city{Harbin}
   \country{China}
 }

\author{Fang Sun}
\authornotemark[1]
\authornote{Corresponding author.}
\email{fsun@cnu.edu.cn}
\orcid{0000-0002-7254-9807}
\affiliation{%
\institution{Capital Normal University}
\city{Beijing}
\country{China}
}


\begin{abstract}
State Space Models (SSMs), particularly Mamba, have shown potential in long-term time series forecasting. However, existing Mamba-based architectures often struggle with datasets characterized by non-stationary patterns. A key observation from time series theory is that the statistical nature of inter-variable relationships differs fundamentally between the trend and seasonal components of a decomposed series. Trend relationships are often driven by a few common stochastic factors or long-run equilibria, suggesting that they reside on a lower-dimensional manifold. In contrast, seasonal relationships involve dynamic, high-dimensional interactions like phase shifts and amplitude co-movements, requiring more expressive modeling. In this paper, we propose DMamba, a novel forecasting model that explicitly aligns architectural complexity with this component-specific characteristic. DMamba employs seasonal-trend decomposition and processes the components with specialized, differentially complex modules: a variable-direction Mamba encoder captures the rich, cross-variable dynamics within the seasonal component, while a simple Multi-Layer Perceptron (MLP) suffices to learn from the lower-dimensional inter-variable relationships in the trend component. Extensive experiments on diverse datasets demonstrate that DMamba sets a new state-of-the-art (SOTA), consistently outperforming both recent Mamba-based architectures and leading decomposition-based models.
\end{abstract}


\maketitle

\section{Introduction}
Long-term time series forecasting (LTSF) is critical for real-world decision-making. 

The technical landscape of LTSF has long been dominated by linear models, such as DLinear\cite{dlinear} and TiDE, and Transformer architectures like Autoformer\cite{autoformer}, FEDformer\cite{fedformer}, and PatchTST\cite{patchtst}. Recently, State Space Models (SSMs)\cite{s4}, spearheaded by the Mamba architecture\cite{mamba}, have emerged as a disruptive frontier. While offering substantial computational efficiency (from $O(L^2)$ to $O(L)$), Mamba’s primary advantage lies in its profound state-space reasoning. Through its content-aware selective scan, Mamba preserves global context more effectively, demonstrating a superior capacity to capture intricate dependencies that are often lost in the diffused attention mechanisms of Transformers. \footnote{Source code and scripts are available at: \url{https://github.com/DMambaKDD/DMamba}}

Following the success of iTransformer\cite{liu2023itransformer}, prominent Mamba TSF models tokenize each variable and use Mamba on the resulting series to model inter-channel dependence\cite{smamba}\cite{bimamba}\cite{ahamed2024timemachine}. While showing promising effectiveness, current Mamba-based models frequently underperform on time series datasets with complicated non-stationarity. For example, SMamba\cite{smamba} lags behind transformer-based models on Electricity Transformer Temperature (ETT) datasets. We hypothesize and validate through experiment that the degradation of performance results from the Mamba module being overwhelmed by the entanglement of multi-scale temporal patterns.

A promising paradigm to address non-stationarity and temporal complexity is to decompose raw time series into semantically distinct components, such as trend and seasonal parts, and apply separate neural networks to process them. Integration of this paradigm into various LTSF architectures, including MLP-based models\cite{rlinear}\cite{dlinear} and transformer-based models \cite{autoformer}\cite{fedformer}, has been successful. This aligns with the statistical insight that the underlying data-generating processes for these components differ significantly. However, existing methods typically employ identical or equally complex modules for both the trend and seasonal component, thus cannot fully exploit this distinction. 

We argue that the nature of inter-variable dependencies within the trend and seasonal components is inherently different, suggesting that they should be modeled with architectures of matching complexity. For example, the phenomenon of cointegration has been observed in multivariate time series data across multiple fields, including economics\cite{engle1987co}\cite{johansen1990maximum}, finance\cite{dwyer1992cointegration}\cite{gatev2006pairs}, energy\cite{siliverstovs2005international}, environmental science\cite{kaufmann2002cointegration} and transportation\cite{inoue2021short}. In cointegrated systems, the joint dynamics of the trend of variables is linearly constrained, thus can often be explained by a few common trends. In contrast, seasonal components encapsulate intricate, periodic fluctuations. The relationships between variables within a seasonal cycle can be highly dynamic, involving shifting phase alignments, amplitude co-movements, and multi-periodic interactions that are not easily captured by simple, low-dimensional representations.


Building upon the above observation, we propose \textbf{DMamba}, a novel multivariate time series forecasting model based on Decomposition and Mamba-based modeling. DMamba's design philosophy is grounded in the principle of architectural parsimony: assign complex model capacity precisely where the data complexity demands it, and use simpler models elsewhere to enhance efficiency and reduce overfitting risks. DMamba utilizes Exponential Moving Average (EMA)\cite{ema} to explicitly decouple the input into stable trend and periodic seasonal components, allowing subsequent modules to specialize. The resulting disentangled streams are then processed through a dual-flow network. A powerful Mamba-based backbone is leveraged specifically to model the purified seasonal residuals, while the trend component is handled by a lightweight MLP. The processed representations from both pathways are integrated and passed through a forecast head to generate the final multi-step predictions.

We conduct comprehensive experiments on a wide range of benchmarks, including the ETT, Weather, and PEMS datasets. Our empirical results demonstrate that \textbf{DMamba} achieves SOTA performance, outperforming both recent Mamba-based models such as S-Mamba and leading decomposition architectures such as TimesNet and XPatch. 
This success is attributed to our hybrid architecture, a conclusion reinforced by extensive ablation studies. These studies demonstrate that DMamba's superior performance stems from its strategic application of a Mamba module to complex seasonal residuals, while retaining a lightweight MLP for stable trend components. This design achieves an excellent balance and robustness, rather than solely benefiting from the robustness of the decomposition method or a specific loss function. Our contributions can be summarized as follows:

\begin{itemize}
    \item \textbf{New Architecture:} We propose \textbf{DMamba}, a simple yet effective decomposition-based forecasting model that, explicitly aligns the complexity of the variable-relation encoder with the intrinsic statistical properties of the trend and seasonal components.
    
    \item \textbf{Theoretical Explanation:} We provide a theoretical motivation for this design from time series theory, arguing that trend relationships are often low-dimensional and thus suitable for simple MLPs, while seasonal relationships require more powerful architectures like Mamba. This insight is empirically validated by our ablation study.
    
    \item \textbf{Comprehensive Benchmarking:} Evaluations on ETT, Weather, and PEMS benchmarks demonstrate a dual breakthrough: DMamba not only sets a SOTA for Mamba-based forecasting on diverse datasets but also represents a Mamba-centric approach to outperform top-tier decomposition architectures like TimesNet and XPatch.
\end{itemize}

\section{Related Work}

Our research draws upon three primary streams of literature: \\Transformer-based forecasting\cite{transformerforecasting}, the emergence of State Space Models (SSMs)\cite{s4} in time series, and decomposition techniques.

\subsection{Transformer-based LTSF Models}
Transformers have become the dominant architecture for Long-term Time Series Forecasting (LTSF) due to their ability to model global dependencies. Early iterations focused on mitigating the $\mathcal{O}(L^2)$ complexity of canonical self-attention; \textbf{Informer}\cite{ett} introduced ProbSparse attention to select significant queries, while \textbf{TimesNet}\cite{timesnet} utilized Fourier transforms to model temporal variations in a 2D space. More recently, the focus has shifted from point-wise attention to subseries-level modeling. \textbf{PatchTST}\cite{patchtst} revolutionized this approach by segmenting time series into patches, enabling the model to capture local semantic information while preserving channel independence. While these architectures have achieved state-of-the-art results, they often require substantial computational resources and may struggle to distinguish essential temporal dependencies from noise in extremely long sequences.

\subsection{Mamba for Time Series Forecasting}
Recently, State Space Models (SSMs), particularly the \textbf{Mamba}\cite{mamba} architecture, have emerged as a powerful alternative to Transformers, offering linear computational complexity $\mathcal{O}(L)$ and a hardware-aware selective scan mechanism. In the domain of time series, \textbf{S-Mamba}\cite{smamba} has leveraged Mamba to capture long-range dependencies with significantly lower memory footprints than Transformers. Counterintuitively, S-Mamba models the inter-variable relationships rather than the inter-temporal relationships using the Mamba architecture, and has demonstrated superior effectiveness of the former through ablation studies. Other works, such as \textbf{Simba}\cite{patro2024simba}, demonstrate Mamba's stability in mixing channel and temporal information. However, despite their efficiency, these singular Mamba-based blocks often face challenges with non-stationary data (e.g., ETT datasets\cite{ett}). The ``entanglement'' of smooth trends and high-frequency fluctuations within the hidden states can degrade the selective scan's ability to isolate stable dynamics, leading to performance degradation—a limitation our work seeks to address.

\subsection{Decomposition and Exponential Smoothing}
To mitigate non-stationarity, series decomposition has become a standard paradigm in LTSF. \textbf{Autoformer}\cite{autoformer} and \textbf{FEDformer}\cite{fedformer} integrated decomposition blocks directly into Transformer layers, while \textbf{DLinear}\cite{dlinear} proved that a simple linear layer applied to decomposed Trend and Seasonal components could outperform complex architectures. These methods typically rely on moving average pooling, which necessitates boundary padding and introduces distributional bias. Recent works have adopted exponential smoothing. \textbf{ETSformer}\cite{etsformer} incorporates Exponential Smoothing Attention (ESA) to replace standard self-attention, while \textbf{CARD}\cite{card} applies exponential smoothing specifically to query and key tokens within the attention head to align temporal distributions. Unlike these Transformer-centric approaches, \textbf{xPatch}\cite{xpatch} utilizes Exponential Moving Average (EMA)\cite{ema} as a pure decomposition module to separate trend and seasonality without padding.

In these decomposition-based methods, the trend and seasonal component are processed by models of similar complexity. A cohesive framework that explicitly aligns model capacity with the intrinsic complexity of inter-variable relationships in decomposed components—applying a powerful variable-relation encoder only where it is truly needed—remains underexplored.

\section{Preliminaries}
\label{sec:preliminaries}

\subsection{Problem Definition}
\label{sec:problem_def}
We focus on the Long-term Time Series Forecasting (LTSF) task. Let $\mathcal{X} = \{\mathbf{x}_1, \dots, \mathbf{x}_L\} \in \mathbb{R}^{L \times D}$ denote the historical look-back window, where $L$ is the input sequence length and $D$ represents the number of variates. The objective is to learn a mapping function $\mathcal{F}: \mathcal{X} \to \mathcal{Y}$ to forecast the future sequence $\mathcal{Y} = \{\mathbf{x}_{L+1}, \dots, \mathbf{x}_{L+T}\} \in \mathbb{R}^{T \times D}$, where $T$ is the prediction horizon. 

\subsection{State Space Models}
\label{sec:ssm_prelim}
The backbone of our proposed architecture leverages the State Space Model (SSM)\cite{s4}, specifically the structured SSM used in Mamba. 
Conceptually, a continuous-time SSM maps a 1-D input function or sequence $x(t) \in \mathbb{R}$ to an output $y(t) \in \mathbb{R}$ via a latent state $h(t) \in \mathbb{R}^N$. The system dynamics are governed by the following linear Ordinary Differential Equation (ODE):
\begin{equation}
    h'(t) = \mathbf{A}h(t) + \mathbf{B}x(t), \quad y(t) = \mathbf{C}h(t)
    \label{eq:continuous_ssm}
\end{equation}
where $\mathbf{A} \in \mathbb{R}^{N \times N}$ is the state evolution matrix\cite{gu2020hippo}, and $\mathbf{B} \in \mathbb{R}^{N \times 1}, \mathbf{C} \in \mathbb{R}^{1 \times N}$ are projection parameters.

To process discrete sequence data, the continuous system in Eq. \eqref{eq:continuous_ssm} is discretized\cite{Discretization}. Adopting the Zero-Order Hold (ZOH) rule with a timescale parameter $\Delta$, the discrete parameters $(\overline{\mathbf{A}}, \overline{\mathbf{B}})$ are derived as:
\begin{equation}
    \overline{\mathbf{A}} = \exp(\Delta \mathbf{A}), \quad \overline{\mathbf{B}} = (\Delta \mathbf{A})^{-1}(\exp(\Delta \mathbf{A}) - \mathbf{I}) \cdot \Delta \mathbf{B}
\end{equation}
Consequently, the discretized state equation becomes a recurrence:
\begin{equation}
    h_t = \overline{\mathbf{A}} h_{t-1} + \overline{\mathbf{B}} x_t, \quad y_t = \mathbf{C} h_t
    \label{eq:discrete_ssm}
\end{equation}
While Eq. \eqref{eq:discrete_ssm} suggests recursive inference, modern implementations (like Mamba) utilize a hardware-aware parallel selective scan algorithm to compute this efficiently with linear complexity $\mathcal{O}(L)$, serving as the fundamental building block for our seasonal modeling. 

\section{Methodology} 
\label{sec:methodology} 


\begin{figure*}[t] 
  \centering 
  \begin{tikzpicture}[ 
      node distance=0.6cm and 0.8cm, 
      cell/.style={rectangle, draw=black!80, thick, minimum width=0.4cm, minimum height=0.4cm, fill=gray!10}, 
      highlight/.style={fill=orange!20, draw=orange!80!black}, 
      box/.style={rectangle, draw=black!80, thick, minimum width=1.4cm, minimum height=0.7cm, fill=white, rounded corners=2pt, font=\footnotesize\bfseries, blur shadow={shadow blur steps=3}, align=center}, 
      module/.style={rectangle, draw=blue!80!black, thick, fill=blue!5, minimum width=1.8cm, minimum height=0.8cm, font=\footnotesize\bfseries, rounded corners=3pt, align=center}, 
      module2/.style={rectangle, color=orange!80!black, thick, fill=blue!5, minimum width=1.8cm, minimum height=0.8cm, font=\footnotesize\bfseries, rounded corners=3pt, align=center}, 
      arrow/.style={-{Stealth[scale=1.0]}, thick, draw=black!80}, 
      text_label/.style={font=\scriptsize\bfseries\itshape, color=black!80, align=center},
      plot_box/.style={rectangle, draw=gray!40, thin, fill=white, minimum width=1.0cm, minimum height=0.5cm, inner sep=0pt}
  ] 
  
      \node[cell, highlight] (in_t) at (0,0) {$\mathbf{X}$}; 
      \node[cell, left=-0.1pt of in_t] (in_2) {}; 
      \node[cell, left=-0.1pt of in_2] (in_1) {}; 
      \node[text_label, above=0.15cm of in_2] {Input}; 
      
      \node[module, right=0.6cm of in_t, fill=green!10, draw=green!60!black] (revin) {RevIN}; 
      \node[module, right=0.6cm of revin, fill=orange!10, draw=orange!60!black] (decomp) {EMA\\Decomp}; 
  
      \draw[arrow] (in_t.east) -- (revin.west); 
      \draw[arrow] (revin.east) -- (decomp.west); 
      
       \begin{scope}[shift={(decomp.south)}, yshift=-0.4cm]
           \draw[->, black!80, -{Stealth[scale=0.5]}] (-0.7,0) -- (0.7,0);
           \draw[->, black!80, -{Stealth[scale=0.5]}] (0,-0.25) -- (0,0.25);
           \draw[black!80!black, thick, smooth] plot[domain=-0.6:0.6, samples=30] (\x, {0.06*sin(30*\x r) + 0.03*sin(60*\x r) + 0.18*\x});
       \end{scope}
   
       \coordinate (split) at ($(decomp.east) + (0.4,0)$); 
       \draw[thick, black!80] (decomp.east) -- (split); 
   
       \node[module, above right=0.4cm and 0.6cm of split] (smamba_emb) {Inverted\\Embed}; 
       \node[box, right=0.5cm of smamba_emb, fill=blue!10, minimum width=2.4cm, minimum height=0.8cm] (smamba_core) {Mamba\\Block}; 
       \node[module, right=0.5cm of smamba_core] (smamba_proj) {Projection}; 
       
       \begin{scope}[shift={(smamba_emb.south)}, yshift=-0.4cm]
           \draw[->, black!80, -{Stealth[scale=0.5]}] (-0.7,0) -- (0.7,0);
           \draw[->, black!80, -{Stealth[scale=0.5]}] (0,-0.25) -- (0,0.25);
           \draw[blue!80!black, thick, smooth] plot[domain=-0.60:0.60, samples=40] (\x, {0.08*sin(35*\x r) + 0.04*cos(70*\x r)});
       \end{scope}
   
       \node[box, below right=0.4cm and 3cm of split, fill=orange!5, minimum width=2.4cm, minimum height=0.8cm] (trend_mlp) {  MLP Block  }; 
       \node[module, right=0.5cm of trend_mlp, fill=orange!5] (trend_proj) {Projection}; 
       
       \begin{scope}[shift={(trend_mlp.south)}, yshift=-0.4cm]
           \draw[->, black!80, -{Stealth[scale=0.5]}] (-0.7,0) -- (0.7,0);
           \draw[->, black!80, -{Stealth[scale=0.5]}] (0,-0.25) -- (0,0.25);
           \draw[orange!80!black, thick, smooth] plot[domain=-0.60:0.60, samples=20] (\x, {0.2 * (1 / (1 + exp(-8*\x)) - 0.5)});
       \end{scope}
  
      \draw[arrow] (split) |- (smamba_emb.west); 
      \draw[arrow] (split) |- (trend_mlp.west); 
      \draw[arrow] (smamba_emb.east) -- (smamba_core.west); 
      \draw[arrow] (smamba_core.east) -- (smamba_proj.west); 
      \draw[arrow] (trend_mlp.east) -- (trend_proj.west); 
  
      \coordinate (merge_v) at ($(smamba_proj.east) + (0.5, -1.8)$); 
      
      \node[module, below=1.0cm of trend_proj, fill=orange!10, draw=orange!60!black] (fusion) {Fusion}; 
      \node[module, left=0.8cm of fusion, fill=green!10, draw=green!60!black] (inv_revin) {Inv RevIN}; 
  
      \node[cell, highlight, left=1.50cm of inv_revin] (out_t) {${\mathbf{Y}}$}; 
      \node[cell, right=-0.1pt of out_t] (out_1) {}; 
      \node[cell, right=-0.1pt of out_1] (out_2) {}; 
      \node[text_label, above=0.15cm of out_1] {Output}; 
  
      \draw[arrow] (smamba_proj.east) -| (merge_v) |- (fusion.east); 
      \draw[arrow] (trend_proj.east) -| (merge_v) |- (fusion.east); 
      
      \draw[arrow] (fusion.west) -- (inv_revin.east); 
      \draw[arrow] (inv_revin.west) -- (out_2.east); 
  
      \node[text_label, color=blue!80!black, above=0.1cm of smamba_core] {Seasonal Stream }; 
      \node[text_label, color=orange!80!black, above=0.1cm of trend_mlp] {Trend Stream }; 
  
  \end{tikzpicture} 
  \caption{The Architecture of DMamba.} 
  \label{fig:architecture} 
  \end{figure*}
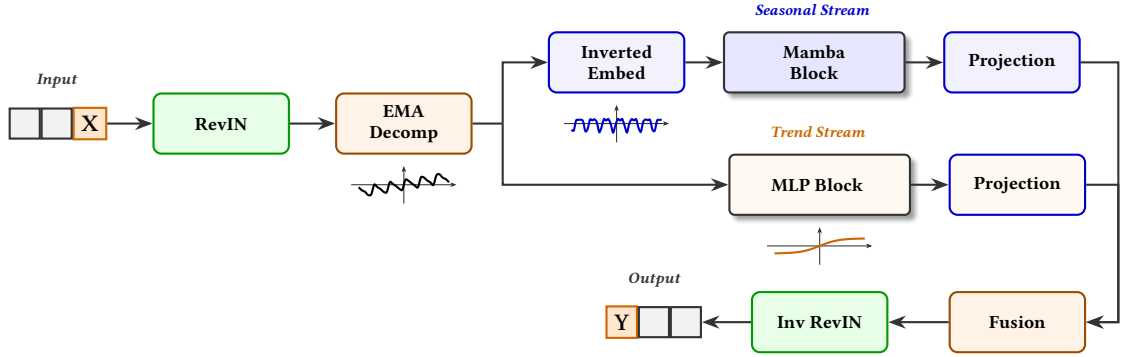

In this section, we present the proposed \textbf{DMamba}. As shown in Figure \ref{fig:architecture}, the architecture is designed to handle non-stationarity through a decouple-and-conquer strategy. The overall workflow proceeds in four stages: (1) \textbf{Normalization} via RevIN\cite{revin} to mitigate distribution shifts; (2) \textbf{Decomposition} using an Exponential Moving Average (EMA)\cite{ema} mechanism to separate trend and seasonality; (3) \textbf{Dual-flow Modeling}, where the Mamba backbone models the seasonal component while an MLP-based head models the trend; and (4) \textbf{Aggregation and Denormalization} to produce the final forecast. 

\subsection{Reversible Instance Normalization} 
Non-stationary time series often suffer from distribution shifts where the mean and variance change over time. To address this, we apply Reversible Instance Normalization (RevIN) before the mixing layers. 
Given the input batch $\mathbf{X} \in \mathbb{R}^{B \times L \times C}$, we compute the instance-specific mean and standard deviation for each channel: 
\begin{equation} 
    \mu = \frac{1}{L} \sum_{i=1}^{L} \mathbf{x}_i, \quad \sigma = \sqrt{\frac{1}{L} \sum_{i=1}^{L} (\mathbf{x}_i - \mu)^2 + \epsilon} 
\end{equation} 
The input is then normalized to $\mathbf{X}'$: 
\begin{equation} 
    \mathbf{X}' = \gamma \left( \frac{\mathbf{X} - \mu}{\sigma} \right) + \beta 
\end{equation} 
where $\gamma, \beta \in \mathbb{R}^C$ are learnable affine parameters. $\mu$ and $\sigma$ are stored to denormalize the output later. 

\subsection{EMA-based Decomposition} 
The normalized input is decomposed into a \textbf{trend component} and a \textbf{seasonal component}. The trend component reflects the long-term progression and the underlying low-frequency direction of the series, effectively smoothing out short-term fluctuations. The seasonal component captures periodic patterns, high-frequency variations, and residual noise that deviate from the long-term trend.

Standard decomposition methods typically use a moving average kernel which introduces "padding bias" at boundaries. To avoid this issue, we implement an Exponential Moving Average (EMA) decomposition module. 
Let $\mathbf{X}'$ be the normalized input. The trend component $\mathbf{X}_{trend} \in \mathbb{R}^{L \times D}$ is computed as: 
\begin{equation} 
    \mathbf{X}_{trend}[t] = \alpha \mathbf{X}'[t] + (1 - \alpha) \mathbf{X}_{trend}[t-1] 
\end{equation} 
The smoothing factor $\alpha$ is a fixed hyperparameter, and for all our experiments, we set $\alpha=0.3$. The seasonal component is obtained by: 
\begin{equation} 
    \mathbf{X}_{seasonal} = \mathbf{X}' - \mathbf{X}_{trend} 
\end{equation}

\subsection{Dual-stream Network} 
The model processes the two components via separate streams to handle their distinct temporal and spatial characteristics. 
\subsubsection{Seasonal Stream (Mamba Backbone)} 
The seasonal component $\mathbf{X}_{seasonal} \in \mathbb{R}^{L \times D}$ is processed by the \textbf{Mamba} backbone. Unlike traditional temporal models that scan time steps, Mamba treats each variate as a token to explicitly model inter-variate dependencies.

\textbf{Variate-wise Inverted Embedding:} For each variate $d \in \{1, \dots, D\}$, the entire sequence of length $L$ is projected into a hidden space of dimension $d_{model}$:
\begin{equation}
    \mathbf{E} = \text{Embed}(\mathbf{X}_{seasonal}) \in \mathbb{R}^{d_{model} \times D}
\end{equation}
This inverted embedding allows the model to treat the $D$ variates as a sequence of tokens.

\begin{figure*}[ht] 
   \centering 
   \resizebox{\linewidth}{!}{
   \begin{tikzpicture}[ 
       module/.style={rectangle, draw=blue!80!black, thick, fill=blue!5, minimum width=1.6cm, minimum height=0.7cm, font=\scriptsize\bfseries, rounded corners=2pt, align=center, blur shadow={shadow blur steps=2}}, 
       norm/.style={rectangle, draw=green!60!black, thick, fill=green!5, minimum width=1.4cm, minimum height=0.6cm, font=\scriptsize\bfseries, rounded corners=2pt, align=center, blur shadow={shadow blur steps=2}}, 
       ffn_box/.style={rectangle, draw=orange!70!black, thick, fill=orange!5, minimum width=2.0cm, minimum height=0.9cm, font=\scriptsize\bfseries, rounded corners=2pt, align=center, blur shadow={shadow blur steps=2}}, 
       flip/.style={rectangle, draw=gray!60, fill=gray!5, minimum width=0.8cm, minimum height=0.45cm, font=\tiny\bfseries, rounded corners=1pt}, 
       op/.style={circle, draw=black!80, thick, fill=white, inner sep=0pt, font=\scriptsize\bfseries, minimum size=0.45cm}, 
       node_text/.style={font=\scriptsize\bfseries, align=center}, 
       arrow/.style={-{Stealth[scale=0.9]}, thick, draw=black!80}, 
       data_label/.style={font=\tiny\itshape, color=blue!70!black} 
   ] 
 
       \node[node_text] (input) at (0,0) {Input $\mathbf{E}$}; 
       \coordinate (split) at (1.2, 0); 
       \draw[thick, black!80] (input.east) -- (split); 
 
       \node[module] (mamba_fwd) at (4.0, 0.6) {$\text{Mamba}_{fwd}$}; 
       
       \node[flip] (flip1) at (2.6, -0.6) {Flip}; 
       \node[module] (mamba_bwd) at (4.0, -0.6) {$\text{Mamba}_{bwd}$}; 
       \node[flip] (flip2) at (5.4, -0.6) {Flip}; 
 
       \draw[arrow] (split) |- (mamba_fwd.west); 
       \draw[arrow] (split) |- (flip1.west); 
       \draw[arrow] (flip1.east) -- (mamba_bwd.west); 
       \draw[arrow] (mamba_bwd.east) -- (flip2.west); 
 
       \node[op] (sum1) at (7.2, 0) {$+$}; 
       \draw[arrow] (mamba_fwd.east) -| (sum1.north) node[pos=0.4, data_label, above] {$\mathbf{H}_{fwd}$}; 
       \draw[arrow] (flip2.east) -| (sum1.south) node[pos=0.4, data_label, below] {$\mathbf{H}_{bwd}$}; 
 
       \draw[thick, black!80] (0.6, 0) -- (0.6, 1.4) -- (7.2, 1.4); 
       \draw[arrow] (7.2, 1.4) -- (sum1.north); 
 
       \node[norm] (ln1) at (8.8, 0) {LayerNorm}; 
       \draw[arrow] (sum1.east) -- (ln1.west); 
 
       \node[ffn_box] (ffn) at (11.0, 0) {FFN \\ \tiny (Conv1d + $\sigma$)}; 
       \draw[arrow] (ln1.east) -- (ffn.west); 
 
       \node[op] (sum2) at (13.0, 0) {$+$}; 
       \draw[arrow] (ffn.east) -- (sum2.west); 
       
       \draw[thick, black!80] (9.6, 0) -- (9.6, 0.8) -- (13.0, 0.8); 
       \draw[arrow] (13.0, 0.8) -- (sum2.north); 
 
       \node[norm] (ln2) at (14.6, 0) {LayerNorm}; 
       \draw[arrow] (sum2.east) -- (ln2.west); 
 
       \node[module, fill=blue!15] (proj) at (16.8, 0) {Projection}; 
       \draw[arrow] (ln2.east) -- (proj.west); 
 
       \node[node_text] (output) at (19.0, 0) {Output $\mathbf{Y}_{seasonal}$}; 
       \draw[arrow] (proj.east) -- (output.west); 
 
       \draw[dashed, gray!60, thick, rounded corners=4pt] 
         (1.8, 1.1) rectangle (6.2, -1.4); 
       \node[font=\scriptsize\bfseries, color=gray!80] at (4.0, 1.6) {Bidirectional SSM Scan}; 
       
       \node[data_label] at (8.8, 0.5) {$\mathbf{H}_{fused}$}; 
 
   \end{tikzpicture} 
   }
   \caption{The Architecture of the Mamba Block.} 
   \label{fig:smamba_final_v2} 
\end{figure*}
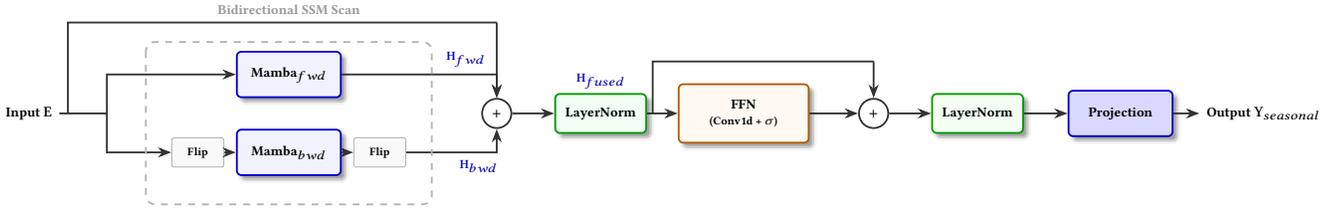

\textbf{Mamba Block:} To capture inter-channel correlations while mitigating causality introduced by ordering the variables, we apply Mamba with both forward and backward directions. As shown in Figure \ref{fig:smamba_final_v2}, each layer in the backbone consists of a forward Mamba block ($\text{Mamba}_{fwd}$) and a backward Mamba block ($\text{Mamba}_{bwd}$). 

The output of the bidirectional scan is fused via a residual connection:
\begin{equation}
    \mathbf{H}_{fwd} = \text{Mamba}_{fwd}(\mathbf{E}), \quad \mathbf{H}_{bwd} = \text{flip}(\text{Mamba}_{bwd}(\text{flip}(\mathbf{E})))
\end{equation}
\begin{equation}
    \mathbf{H}_{fused} = \text{LayerNorm}(\mathbf{E} + \mathbf{H}_{fwd} + \mathbf{H}_{bwd})
\end{equation}

Subsequently, a Feed-Forward Network (FFN), consisting of two $1 \times 1$ convolutions and a non-linear activation $\sigma$, is applied:
\begin{equation}
    \mathbf{H}_{out} = \text{LayerNorm}(\mathbf{H}_{fused} + \text{Dropout}(\text{Conv1d}(\sigma(\text{Conv1d}(\mathbf{H}_{fused})))))
\end{equation}
Finally, the representation is projected to the prediction horizon $T$:
\begin{equation} 
    \mathbf{Y}_{seasonal} = \text{Project}(\mathbf{H}_{out}) \in \mathbb{R}^{T \times D} 
\end{equation} 

\begin{table*}[t]
  \caption{We compare DMamba with 10 competitive baselines. The look-back length $L$ is set to 96 and the forecast length $T$ is set to $\{96, 192, 336, 720\}$. Best results are in \textbf{bold}, and the second best are \underline{underlined}.}
  \label{tab:results_main}
  \centering
  \renewcommand{\arraystretch}{0.9} 
  \setlength{\tabcolsep}{0.8pt}
  \tiny
  \resizebox{\textwidth}{!}{
  \begin{tabular}{lc *{11}{cc}}
    \toprule
    \multicolumn{2}{c}{Models} & \multicolumn{2}{c}{\textbf{DMamba}} & \multicolumn{2}{c}{S-Mamba} & \multicolumn{2}{c}{TimePro$^\ast$} & \multicolumn{2}{c}{iTransformer} & \multicolumn{2}{c}{PatchTST} & \multicolumn{2}{c}{RLinear} & \multicolumn{2}{c}{DLinear} & \multicolumn{2}{c}{TiDE} & \multicolumn{2}{c}{FEDformer} & \multicolumn{2}{c}{TimesNet} & \multicolumn{2}{c}{XPatch} \\
    \cmidrule(lr){3-4} \cmidrule(lr){5-6} \cmidrule(lr){7-8} \cmidrule(lr){9-10} \cmidrule(lr){11-12} \cmidrule(lr){13-14} \cmidrule(lr){15-16} \cmidrule(lr){17-18} \cmidrule(lr){19-20} \cmidrule(lr){21-22} \cmidrule(lr){23-24}
    Metric & $T$ & MSE & MAE & MSE & MAE & MSE & MAE & MSE & MAE & MSE & MAE & MSE & MAE & MSE & MAE & MSE & MAE & MSE & MAE & MSE & MAE & MSE & MAE \\ 
    \midrule
    \multirow{5}{*}{\rotatebox{90}{ETTh1}} 
     & 96  & \underline{0.377} & \textbf{0.390} & 0.386 & 0.405 & -- & -- & 0.386 & 0.405 & 0.414 & 0.419 & 0.386 & 0.395 & 0.386 & 0.400 & 0.479 & 0.464 & \textbf{0.376} & 0.419 & 0.384 & 0.402 & 0.425 & 0.427 \\
     & 192 & \textbf{0.414} & \textbf{0.407} & 0.443 & 0.437 & -- & -- & 0.441 & 0.436 & 0.460 & 0.445 & 0.437 & \underline{0.424} & 0.437 & 0.432 & 0.525 & 0.492 & \underline{0.420} & 0.448 & 0.436 & 0.429 & 0.436 & 0.425 \\
     & 336 & \textbf{0.441} & \textbf{0.427} & 0.489 & 0.468 & -- & -- & 0.487 & \underline{0.458} & 0.501 & 0.466 & 0.479 & 0.446 & 0.481 & 0.459 & 0.565 & 0.515 & \underline{0.459} & 0.465 & 0.491 & 0.469 & 0.454 & 0.433 \\
     & 720 & \textbf{0.465} & \textbf{0.457} & 0.502 & 0.489 & -- & -- & 0.503 & 0.491 & 0.500 & 0.488 & 0.481 & 0.470 & 0.519 & 0.516 & 0.594 & 0.558 & 0.506 & 0.507 & 0.521 & 0.500 & \underline{0.478} & \underline{0.461} \\ 
     \cmidrule{2-24}
     & Avg & \textbf{0.424} & \textbf{0.420} & 0.455 & 0.450 & \underline{0.438} & \underline{0.438} & 0.454 & 0.447 & 0.469 & 0.454 & 0.446 & 0.434 & 0.456 & 0.452 & 0.541 & 0.507 & 0.440 & 0.460 & 0.458 & 0.450 & 0.448 & 0.437 \\ \midrule
    \multirow{5}{*}{\rotatebox{90}{ETTh2}} 
     & 96  & \textbf{0.235} & \textbf{0.301} & 0.296 & 0.348 & -- & -- & 0.297 & 0.349 & 0.302 & 0.348 & 0.288 & 0.338 & 0.333 & 0.387 & 0.400 & 0.440 & 0.358 & 0.397 & 0.340 & 0.374 & \underline{0.237} & \underline{0.307} \\
     & 192 & \textbf{0.291} & \textbf{0.335} & 0.376 & 0.396 & -- & -- & 0.380 & 0.400 & 0.388 & 0.400 & \underline{0.374} & \underline{0.390} & 0.477 & 0.476 & 0.528 & 0.509 & 0.429 & 0.439 & 0.402 & 0.414 & \underline{0.293} & 0.341 \\
     & 336 & \underline{0.342} & \textbf{0.374} & 0.424 & 0.431 & -- & -- & 0.428 & 0.432 & 0.426 & 0.433 & \textbf{0.415} & \underline{0.426} & 0.594 & 0.541 & 0.643 & 0.571 & 0.496 & 0.487 & 0.452 & 0.452 & \textbf{0.340} & \underline{0.376} \\
     & 720 & \underline{0.404} & \textbf{0.425} & 0.426 & 0.444 & -- & -- & 0.427 & 0.445 & 0.431 & 0.446 & \textbf{0.420} & \underline{0.440} & 0.831 & 0.657 & 0.874 & 0.679 & 0.463 & 0.474 & 0.462 & 0.468 & \textbf{0.403} & \underline{0.426} \\ 
     \cmidrule{2-24}
     & Avg & \textbf{0.318} & \textbf{0.359} & 0.381 & 0.405 & \underline{0.377} & \underline{0.403} & 0.383 & 0.407 & 0.387 & 0.407 & 0.374 & 0.398 & 0.559 & 0.515 & 0.611 & 0.550 & 0.437 & 0.449 & 0.414 & 0.427 & \textbf{0.318} & 0.363 \\ \midrule
    \multirow{5}{*}{\rotatebox{90}{ETTm1}} 
     & 96  & \textbf{0.307} & \textbf{0.345} & 0.333 & 0.368 & -- & -- & 0.334 & 0.368 & \underline{0.329} & \underline{0.367} & 0.355 & 0.376 & 0.345 & 0.372 & 0.364 & 0.387 & 0.379 & 0.419 & 0.338 & 0.375 & \underline{0.323} & \underline{0.354} \\
     & 192 & \textbf{0.350} & \textbf{0.372} & 0.376 & 0.390 & -- & -- & 0.377 & 0.391 & \underline{0.367} & \underline{0.385} & 0.391 & 0.392 & 0.380 & 0.389 & 0.398 & 0.404 & 0.426 & 0.441 & 0.374 & 0.387 & \underline{0.356} & \underline{0.372} \\
     & 336 & \textbf{0.385} & \textbf{0.391} & 0.408 & 0.413 & -- & -- & 0.426 & 0.420 & \underline{0.399} & \underline{0.410} & 0.424 & \underline{0.415} & 0.413 & 0.413 & 0.428 & 0.425 & 0.445 & 0.459 & 0.410 & 0.411 & \underline{0.394} & \underline{0.395} \\
     & 720 & \underline{0.462} & \underline{0.430} & 0.475 & 0.448 & -- & -- & 0.491 & 0.459 & \textbf{0.454} & \underline{0.439} & 0.487 & 0.450 & \underline{0.474} & 0.453 & 0.487 & 0.461 & 0.543 & 0.490 & 0.478 & 0.450 & 0.465 & \textbf{0.430} \\ 
     \cmidrule{2-24}
     & Avg & \textbf{0.376} & \textbf{0.385} & 0.398 & 0.405 & 0.391 & \underline{0.400} & 0.407 & 0.410 & \underline{0.387} & \underline{0.400} & 0.414 & 0.407 & 0.403 & 0.407 & 0.419 & 0.419 & 0.448 & 0.452 & 0.400 & 0.406 & \underline{0.385} & \underline{0.388} \\ \midrule
    \multirow{5}{*}{\rotatebox{90}{ETTm2}} 
     & 96  & \textbf{0.164} & \textbf{0.247} & 0.179 & 0.263 & -- & -- & 0.180 & 0.264 & 0.175 & 0.259 & 0.182 & 0.265 & 0.193 & 0.292 & 0.207 & 0.305 & 0.203 & 0.287 & 0.187 & 0.267 & \underline{0.167} & \underline{0.250} \\
     & 192 & \textbf{0.229} & \textbf{0.291} & 0.250 & 0.309 & -- & -- & 0.250 & 0.309 & 0.241 & \underline{0.302} & \underline{0.246} & 0.304 & 0.284 & 0.362 & 0.290 & 0.364 & 0.269 & 0.328 & 0.249 & 0.309 & \underline{0.231} & \underline{0.292} \\
     & 336 & \textbf{0.292} & \textbf{0.331} & 0.312 & 0.349 & -- & -- & 0.311 & 0.348 & 0.305 & 0.343 & \underline{0.307} & \underline{0.342} & 0.369 & 0.427 & 0.377 & 0.422 & 0.325 & 0.366 & 0.321 & 0.351 & \textbf{0.292} & \underline{0.332} \\
     & 720 & \textbf{0.380} & \textbf{0.384} & 0.411 & 0.406 & -- & -- & 0.412 & 0.407 & \underline{0.402} & \underline{0.400} & 0.407 & \underline{0.398} & 0.554 & 0.522 & 0.558 & 0.524 & 0.421 & 0.415 & 0.408 & 0.403 & \textbf{0.380} & \underline{0.385} \\ 
     \cmidrule{2-24}
     & Avg & \textbf{0.266} & \textbf{0.313} & 0.288 & 0.332 & \underline{0.281} & \underline{0.326} & 0.288 & 0.332 & \underline{0.281} & \underline{0.326} & 0.286 & \underline{0.327} & 0.350 & 0.401 & 0.358 & 0.404 & 0.305 & 0.349 & 0.291 & 0.333 & \underline{0.267} & \underline{0.314} \\ \midrule
    \multirow{5}{*}{\rotatebox{90}{Elec.}} 
     & 96  & \underline{0.144} & \textbf{0.233} & \textbf{0.139} & \underline{0.235} & -- & -- & \underline{0.148} & 0.240 & 0.181 & 0.270 & 0.201 & 0.281 & 0.197 & 0.282 & 0.237 & 0.329 & 0.193 & 0.308 & 0.168 & 0.272 & 0.192 & 0.267 \\
     & 192 & \textbf{0.156} & \textbf{0.245} & \underline{0.159} & 0.255 & -- & -- & 0.162 & \underline{0.253} & 0.188 & 0.274 & 0.201 & 0.283 & 0.196 & 0.285 & 0.236 & 0.330 & 0.201 & 0.315 & 0.184 & 0.289 & 0.190 & 0.269 \\
     & 336 & \textbf{0.172} & \textbf{0.261} & \underline{0.176} & 0.272 & -- & -- & 0.178 & \underline{0.269} & 0.204 & 0.293 & 0.215 & 0.298 & 0.209 & 0.301 & 0.249 & 0.344 & 0.214 & 0.329 & 0.198 & 0.300 & 0.206 & 0.284 \\
     & 720 & \underline{0.206} & \textbf{0.293} & \textbf{0.204} & \underline{0.298} & -- & -- & 0.225 & 0.317 & 0.246 & 0.324 & 0.257 & 0.331 & 0.245 & 0.333 & 0.284 & 0.373 & 0.246 & 0.355 & \underline{0.220} & 0.320 & 0.244 & 0.314 \\ 
     \cmidrule{2-24}
     & Avg & \underline{0.170} & \underline{0.258} & \underline{0.170} & 0.265 & \textbf{0.169} & \textbf{0.262} & 0.178 & 0.270 & 0.205 & 0.290 & 0.219 & 0.298 & 0.212 & 0.300 & 0.251 & 0.344 & 0.214 & 0.327 & 0.192 & 0.295 & 0.208 & 0.284 \\ \midrule
    \multirow{5}{*}{\rotatebox{90}{Exch.}} 
     & 96  & \underline{0.080} & \textbf{0.198} & \textbf{0.086} & 0.207 & -- & -- & \textbf{0.086} & \underline{0.206} & \underline{0.088} & \underline{0.205} & 0.093 & 0.217 & \underline{0.088} & 0.218 & 0.094 & 0.218 & 0.148 & 0.278 & 0.107 & 0.234 & 0.085 & \underline{0.202} \\
     & 192 & \textbf{0.171} & \textbf{0.294} & 0.182 & 0.304 & -- & -- & \underline{0.177} & \underline{0.299} & \underline{0.176} & \underline{0.299} & 0.184 & 0.307 & \underline{0.176} & 0.315 & 0.184 & 0.307 & 0.271 & 0.315 & 0.226 & 0.344 & \underline{0.177} & \underline{0.298} \\
     & 336 & \underline{0.337} & \underline{0.418} & 0.332 & 0.418 & -- & -- & \underline{0.331} & \underline{0.417} & \textbf{0.301} & \textbf{0.397} & 0.351 & 0.432 & \underline{0.313} & 0.427 & 0.349 & 0.431 & 0.460 & 0.427 & 0.367 & 0.448 & 0.345 & 0.422 \\
     & 720 & \textbf{0.847} & 0.693 & \underline{0.867} & 0.703 & -- & -- & \textbf{0.847} & \underline{0.691} & 0.901 & 0.714 & 0.886 & 0.714 & \underline{0.839} & \textbf{0.695} & 0.852 & 0.698 & 1.195 & \textbf{0.695} & 0.964 & 0.746 & 0.877 & 0.707 \\ 
     \cmidrule{2-24}
     & Avg & \underline{0.359} & \underline{0.401} & 0.367 & 0.408 & \textbf{0.352} & \textbf{0.399} & \underline{0.360} & \underline{0.403} & \underline{0.367} & \underline{0.404} & 0.378 & 0.417 & \underline{0.354} & 0.414 & 0.370 & 0.413 & 0.519 & 0.429 & 0.416 & 0.443 & 0.371 & 0.407 \\ \midrule
    \multirow{5}{*}{\rotatebox{90}{Weath.}} 
     & 96  & \underline{0.166} & \textbf{0.201} & \textbf{0.165} & \underline{0.210} & -- & -- & 0.174 & \underline{0.214} & 0.177 & 0.218 & 0.192 & 0.232 & 0.196 & 0.255 & 0.202 & 0.261 & 0.217 & 0.296 & 0.172 & 0.220 & 0.190 & 0.222 \\ 
     & 192 & \underline{0.216} & \textbf{0.247} & \textbf{0.214} & \underline{0.252} & -- & -- & 0.221 & \underline{0.254} & 0.225 & 0.259 & 0.240 & 0.271 & 0.237 & 0.296 & 0.242 & 0.298 & 0.276 & 0.336 & 0.219 & 0.261 & 0.232 & 0.257 \\ 
     & 336 & \textbf{0.240} & \textbf{0.278} & 0.274 & 0.297 & -- & -- & 0.278 & 0.296 & \underline{0.278} & \underline{0.297} & 0.292 & 0.307 & \underline{0.283} & 0.335 & 0.287 & 0.335 & 0.339 & 0.380 & 0.280 & 0.306 & \underline{0.248} & \underline{0.280} \\ 
     & 720 & \textbf{0.316} & \underline{0.330} & 0.350 & 0.345 & -- & -- & 0.358 & 0.347 & 0.354 & 0.348 & 0.364 & 0.353 & \underline{0.345} & 0.381 & 0.351 & 0.386 & 0.403 & 0.428 & 0.365 & 0.359 & \underline{0.318} & \textbf{0.326} \\ 
     \cmidrule{2-24}
     & Avg & \textbf{0.234} & \textbf{0.264} & \underline{0.251} & \underline{0.276} & \underline{0.251} & \underline{0.276} & 0.258 & 0.278 & 0.259 & 0.282 & 0.272 & 0.291 & 0.265 & 0.317 & 0.271 & 0.320 & 0.309 & 0.360 & 0.259 & 0.287 & \underline{0.247} & \underline{0.271} \\ 
    \bottomrule
  \end{tabular}
  }
  \begin{flushleft}
    \scriptsize $^\ast$ The original paper of TimePro does not provide specific metrics for each forecast length.
  \end{flushleft}
\end{table*}

\begin{table*}[t]
  \caption{Full results on PEMS datasets. We compare DMamba with 9 competitive baselines. The look-back length $L$ is set to 96 and the forecast length $T$ is set to $\{12, 24, 48, 96\}$. Best results are in \textbf{bold}, and the second best are \underline{underlined}.}
  \label{tab:results_pems}
  \centering
  \renewcommand{\arraystretch}{0.9} 
  \setlength{\tabcolsep}{1.1pt} 
  \tiny
  \resizebox{\textwidth}{!}{
  \begin{tabular}{lc *{10}{cc}}
    \toprule
    \multicolumn{2}{c}{Models} & \multicolumn{2}{c}{\textbf{DMamba}} & \multicolumn{2}{c}{S-Mamba} & \multicolumn{2}{c}{iTransformer} & \multicolumn{2}{c}{PatchTST} & \multicolumn{2}{c}{RLinear} & \multicolumn{2}{c}{DLinear} & \multicolumn{2}{c}{TiDE} & \multicolumn{2}{c}{FEDformer} & \multicolumn{2}{c}{TimesNet} & \multicolumn{2}{c}{XPatch} \\
    \cmidrule(lr){3-4} \cmidrule(lr){5-6} \cmidrule(lr){7-8} \cmidrule(lr){9-10} \cmidrule(lr){11-12} \cmidrule(lr){13-14} \cmidrule(lr){15-16} \cmidrule(lr){17-18} \cmidrule(lr){19-20} \cmidrule(lr){21-22}
    Metric & $T$ & MSE & MAE & MSE & MAE & MSE & MAE & MSE & MAE & MSE & MAE & MSE & MAE & MSE & MAE & MSE & MAE & MSE & MAE & MSE & MAE \\ 
    \midrule
    \multirow{5}{*}{\rotatebox{90}{PEMS03}} 
     & 12 & \textbf{0.060} & \textbf{0.161} & \underline{0.065} & \underline{0.169} & 0.071 & 0.174 & 0.099 & 0.216 & 0.126 & 0.236 & 0.122 & 0.243 & 0.178 & 0.305 & 0.126 & 0.251 & 0.085 & 0.192 & 0.090 & 0.197 \\
     & 24 & \textbf{0.077} & \textbf{0.183} & \underline{0.087} & \underline{0.196} & 0.093 & 0.201 & 0.142 & 0.259 & 0.246 & 0.334 & 0.201 & 0.317 & 0.257 & 0.371 & 0.149 & 0.275 & 0.118 & 0.223 & 0.140 & 0.248 \\
     & 48 & \textbf{0.109} & \textbf{0.217} & 0.133 & 0.243 & \underline{0.125} & \underline{0.236} & 0.211 & 0.319 & 0.551 & 0.529 & 0.333 & 0.425 & 0.379 & 0.463 & 0.227 & 0.348 & 0.155 & 0.260 & 0.255 & 0.338 \\
     & 96 & \textbf{0.147} & \textbf{0.256} & 0.201 & 0.305 & \underline{0.164} & \underline{0.275} & 0.269 & 0.370 & 1.057 & 0.787 & 0.457 & 0.515 & 0.490 & 0.539 & 0.348 & 0.434 & 0.228 & 0.317 & 0.380 & 0.423 \\ 
     \cmidrule{2-22}
     & Avg & \textbf{0.098} & \textbf{0.204} & 0.122 & 0.228 & \underline{0.113} & \underline{0.221} & 0.180 & 0.291 & 0.495 & 0.472 & 0.278 & 0.375 & 0.326 & 0.419 & 0.213 & 0.327 & 0.147 & 0.248 & 0.216 & 0.301 \\ \midrule
    \multirow{5}{*}{\rotatebox{90}{PEMS04}} 
     & 12 & \textbf{0.074} & \textbf{0.174} & \underline{0.076} & \underline{0.180} & 0.078 & 0.183 & 0.105 & 0.224 & 0.138 & 0.252 & 0.148 & 0.272 & 0.219 & 0.340 & 0.138 & 0.262 & 0.087 & 0.195 & 0.106 & 0.214 \\
     & 24 & \underline{0.089} & \underline{0.194} & \textbf{0.084} & \textbf{0.193} & 0.095 & 0.205 & 0.153 & 0.275 & 0.258 & 0.348 & 0.224 & 0.340 & 0.292 & 0.398 & 0.177 & 0.293 & 0.103 & 0.215 & 0.168 & 0.274 \\
     & 48 & \textbf{0.113} & \textbf{0.220} & \underline{0.115} & \underline{0.224} & 0.120 & 0.233 & 0.229 & 0.339 & 0.572 & 0.544 & 0.355 & 0.437 & 0.409 & 0.478 & 0.270 & 0.368 & 0.136 & 0.250 & 0.308 & 0.375 \\
     & 96 & \underline{0.143} & \underline{0.251} & \textbf{0.137} & \textbf{0.248} & 0.150 & 0.262 & 0.291 & 0.389 & 1.137 & 0.820 & 0.452 & 0.504 & 0.492 & 0.532 & 0.341 & 0.427 & 0.190 & 0.303 & 0.491 & 0.492 \\ 
     \cmidrule{2-22}
     & Avg & \underline{0.105} & \textbf{0.210} & \textbf{0.103} & \underline{0.211} & 0.111 & 0.221 & 0.195 & 0.307 & 0.526 & 0.491 & 0.295 & 0.388 & 0.353 & 0.437 & 0.231 & 0.337 & 0.129 & 0.241 & 0.268 & 0.339 \\ \midrule
    \multirow{5}{*}{\rotatebox{90}{PEMS07}} 
     & 12 & \textbf{0.059} & \textbf{0.150} & \underline{0.063} & \underline{0.159} & 0.067 & 0.165 & 0.095 & 0.207 & 0.118 & 0.235 & 0.115 & 0.242 & 0.173 & 0.304 & 0.109 & 0.225 & 0.082 & 0.181 & 0.083 & 0.189 \\
     & 24 & \textbf{0.077} & \textbf{0.173} & \underline{0.081} & \underline{0.183} & 0.088 & 0.190 & 0.150 & 0.262 & 0.242 & 0.341 & 0.210 & 0.329 & 0.271 & 0.383 & 0.125 & 0.244 & 0.101 & 0.204 & 0.138 & 0.244 \\
     & 48 & \underline{0.096} & \underline{0.198} & \textbf{0.093} & \textbf{0.192} & 0.110 & 0.215 & 0.253 & 0.340 & 0.562 & 0.541 & 0.398 & 0.458 & 0.446 & 0.495 & 0.165 & 0.288 & 0.134 & 0.238 & 0.265 & 0.342 \\
     & 96 & \underline{0.131} & \underline{0.234} & \textbf{0.117} & \textbf{0.217} & 0.139 & 0.245 & 0.346 & 0.404 & 1.096 & 0.795 & 0.594 & 0.553 & 0.628 & 0.577 & 0.262 & 0.376 & 0.181 & 0.279 & 0.443 & 0.453 \\ 
     \cmidrule{2-22}
     & Avg & \underline{0.091} & \underline{0.189} & \textbf{0.089} & \textbf{0.188} & 0.101 & 0.204 & 0.211 & 0.303 & 0.504 & 0.478 & 0.329 & 0.395 & 0.380 & 0.440 & 0.165 & 0.283 & 0.124 & 0.225 & 0.232 & 0.307 \\ \midrule
    \multirow{5}{*}{\rotatebox{90}{PEMS08}} 
     & 12 & \textbf{0.067} & \textbf{0.163} & \underline{0.076} & \underline{0.178} & 0.079 & 0.182 & 0.168 & 0.232 & 0.133 & 0.247 & 0.154 & 0.276 & 0.227 & 0.343 & 0.173 & 0.273 & 0.112 & 0.212 & 0.097 & 0.204 \\
     & 24 & \textbf{0.080} & \textbf{0.180} & \underline{0.104} & \underline{0.209} & 0.115 & 0.219 & 0.224 & 0.281 & 0.249 & 0.343 & 0.248 & 0.353 & 0.318 & 0.409 & 0.210 & 0.301 & 0.141 & 0.238 & 0.157 & 0.262 \\
     & 48 & \textbf{0.103} & \textbf{0.205} & \underline{0.167} & \underline{0.228} & 0.186 & 0.235 & 0.321 & 0.354 & 0.569 & 0.544 & 0.440 & 0.470 & 0.497 & 0.510 & \underline{0.320} & 0.394 & 0.198 & 0.283 & 0.281 & 0.357 \\
     & 96 & \textbf{0.145} & \textbf{0.248} & \underline{0.245} & \underline{0.280} & \underline{0.221} & \underline{0.267} & 0.408 & 0.417 & 1.166 & 0.814 & 0.674 & 0.565 & 0.721 & 0.592 & 0.442 & 0.465 & 0.320 & 0.351 & 0.487 & 0.477 \\ 
     \cmidrule{2-22}
     & Avg & \textbf{0.099} & \textbf{0.199} & \underline{0.148} & \underline{0.224} & 0.150 & 0.226 & 0.280 & 0.321 & 0.529 & 0.487 & 0.379 & 0.416 & 0.441 & 0.464 & 0.286 & 0.358 & 0.193 & 0.271 & 0.255 & 0.325 \\ 
    \bottomrule
  \end{tabular}
  }
\end{table*}

\begin{table}[h!]
  \caption{Ablation Study on Branch Architectures across ETT Datasets. We evaluate the effectiveness of the Mamba block by placing it in different branches: \textbf{DMamba} (Mamba for Seasonal, MLP for Trend), \textbf{All-MLP} (MLP for both), \textbf{Mamba-Trend} (MLP for Seasonal, Mamba for Trend), \textbf{All-Mamba} (Mamba for both), and \textbf{T-Mamba} (Temporal Mamba for Seasonal). Best results are in \textbf{bold}, and the second best are \underline{underlined}.}
  \label{tab:ablation_final_updated}
  \centering
  \renewcommand{\arraystretch}{1.0} 
  \setlength{\tabcolsep}{1.2pt} 
  \tiny
  \resizebox{\columnwidth}{!}{
  \begin{tabular}{lc cc cc cc cc cc}
    \toprule
    \multicolumn{2}{c}{Models} & \multicolumn{2}{c}{\textbf{DMamba}} & \multicolumn{2}{c}{All-MLP} & \multicolumn{2}{c}{Mamba-Trend} & \multicolumn{2}{c}{All-Mamba} & \multicolumn{2}{c}{T-Mamba} \\
    \cmidrule(lr){3-4} \cmidrule(lr){5-6} \cmidrule(lr){7-8} \cmidrule(lr){9-10} \cmidrule(lr){11-12}
    Dataset & $T$ & MSE & MAE & MSE & MAE & MSE & MAE & MSE & MAE & MSE & MAE \\ 
    \midrule
    \multirow{5}{*}{ETTh1} 
     & 96  & \underline{0.377} & \underline{0.390} & \textbf{0.376} & \textbf{0.392} & 0.379 & 0.398 & 0.377 & 0.394 & 0.380 & 0.396 \\
     & 192 & \textbf{0.414} & \textbf{0.407} & 0.430 & 0.418 & 0.435 & 0.426 & \underline{0.431} & \underline{0.421} & 0.437 & 0.423 \\
     & 336 & \textbf{0.441} & \textbf{0.427} & \underline{0.471} & \underline{0.437} & 0.477 & 0.445 & 0.472 & 0.440 & 0.484 & 0.442 \\
     & 720 & \underline{0.465} & \underline{0.457} & \textbf{0.461} & \textbf{0.459} & 0.482 & 0.467 & 0.471 & 0.461 & 0.498 & 0.468 \\ 
     \cmidrule{2-12}
     & \textbf{Avg} & \textbf{0.424} & \textbf{0.421} & \underline{0.435} & \underline{0.427} & 0.443 & 0.434 & 0.438 & 0.429 & 0.449 & 0.432 \\
    \midrule
    \multirow{5}{*}{ETTh2} 
     & 96  & \textbf{0.235} & \textbf{0.301} & 0.294 & 0.338 & 0.288 & 0.336 & \underline{0.285} & \underline{0.333} & 0.298 & 0.339 \\
     & 192 & \textbf{0.291} & \textbf{0.335} & 0.371 & 0.386 & 0.361 & 0.382 & \underline{0.358} & \underline{0.379} & 0.367 & 0.384 \\
     & 336 & \textbf{0.342} & \textbf{0.374} & 0.379 & 0.402 & \underline{0.371} & \underline{0.397} & 0.374 & 0.399 & 0.380 & 0.403 \\
     & 720 & 0.404 & 0.425 & 0.411 & 0.430 & \underline{0.398} & \underline{0.422} & \textbf{0.393} & \textbf{0.417} & 0.414 & 0.432 \\ \cmidrule{2-12}
     & \textbf{Avg} & \textbf{0.318} & \textbf{0.359} & 0.364 & 0.389 & 0.355 & 0.384 & \underline{0.353} & \underline{0.382} & 0.365 & 0.389 \\ \midrule
    \multirow{5}{*}{ETTm1} 
     & 96  & \textbf{0.307} & \underline{0.345} & 0.327 & 0.348 & 0.326 & 0.352 & 0.319 & \textbf{0.345} & 0.317 & \textbf{0.345} \\
     & 192 & \textbf{0.350} & \textbf{0.372} & 0.374 & \underline{0.371} & 0.369 & 0.374 & 0.369 & 0.372 & 0.371 & 0.374 \\
     & 336 & \textbf{0.385} & \textbf{0.391} & 0.405 & 0.392 & 0.400 & 0.394 & 0.401 & 0.394 & \underline{0.390} & \underline{0.392} \\
     & 720 & \textbf{0.462} & \underline{0.430} & 0.467 & \textbf{0.429} & 0.480 & 0.438 & 0.470 & 0.434 & 0.455 & \textbf{0.429} \\ \cmidrule{2-12}
     & \textbf{Avg} & \textbf{0.376} & \textbf{0.385} & 0.393 & 0.385 & 0.394 & 0.390 & 0.390 & 0.386 & \underline{0.383} & \underline{0.385} \\ \midrule
    \multirow{5}{*}{ETTm2} 
     & 96  & \textbf{0.164} & \textbf{0.247} & 0.179 & 0.255 & 0.173 & 0.252 & \underline{0.172} & \underline{0.252} & 0.175 & 0.253 \\
     & 192 & \textbf{0.229} & \textbf{0.291} & 0.243 & 0.297 & 0.242 & 0.298 & \underline{0.239} & \underline{0.295} & 0.240 & 0.296 \\
     & 336 & \textbf{0.292} & \textbf{0.331} & 0.300 & \underline{0.334} & 0.299 & 0.335 & 0.304 & 0.337 & 0.303 & 0.335 \\
     & 720 & \textbf{0.380} & \textbf{0.384} & \underline{0.395} & \underline{0.391} & 0.400 & 0.394 & 0.398 & 0.392 & 0.400 & 0.393 \\ \cmidrule{2-12}
     & \textbf{Avg} & \textbf{0.266} & \textbf{0.313} & 0.279 & 0.319 & 0.279 & 0.320 & \underline{0.278} & \underline{0.319} & \underline{0.279} & \underline{0.319} \\ 
    \bottomrule
  \end{tabular}
  }
\end{table}

\subsubsection{Trend Stream}
The trend component $\mathbf{X}_{trend} \in \mathbb{R}^{L \times D}$ is modeled via a channel-independent MLP to capture stable low-frequency patterns. We employ a hierarchical structure consisting of $N$ stacked layers:
\begin{equation}
    \mathbf{H}_l = \text{LayerNorm}(\text{AvgPool}(\mathbf{W}_l \mathbf{H}_{l-1})), \quad l=1,\dots,N
\end{equation}
\begin{equation}
    \mathbf{Y}_{trend} = \mathbf{W}_{out} \mathbf{H}_N \in \mathbb{R}^{T \times D}
\end{equation}
where $\mathbf{H}_0 = \mathbf{X}_{trend}$. The pooling operation $\text{AvgPool}(\cdot)$ effectively suppresses high-frequency noise, while the linear projection $\mathbf{W}_{out}$ maps features to the prediction horizon $T$.

\subsection{Aggregation and Denormalization} 
The final forecast $\mathbf{Y}'_{pred}$ is produced by a linear fusion layer following the concatenation of seasonal and trend predictions:
\begin{equation} 
    \mathbf{Y}'_{pred} = \text{Linear}(\text{Concat}(\mathbf{Y}_{seasonal}, \mathbf{Y}_{trend})) \in \mathbb{R}^{T \times D} 
\end{equation} 
We then restore the original scale using the inverse RevIN transformation: 
\begin{equation} 
    \hat{\mathbf{Y}} = \frac{\mathbf{Y}'_{pred} - \beta}{\gamma} \cdot \sigma + \mu 
\end{equation} 

\subsection{Arctan Weighted Loss Function} 
Standard MSE/MAE losses treat all time steps equally, ignoring the increasing uncertainty over longer horizons. We use the \textbf{Arctan Weighted Loss}\cite{card} for training to prioritize near-term accuracy. For each step $t \in \{0, \dots, T-1\}$, the weighting factor $w_t$ is defined as:
\begin{equation} 
    w_t = -\arctan(t+1) + \frac{\pi}{4} + 1 
\end{equation} 
This weight monotonically decays from $\sim$1.0 to $\sim$0.21, focusing optimization on more certain, immediate horizons. The training objective $\mathcal{L}$ computes the weighted absolute error:
\begin{equation} 
    \mathcal{L} = \frac{1}{T \cdot D} \sum_{t=0}^{T-1} \sum_{d=1}^{D} w_t \cdot |\hat{y}_{t,d} - y_{t,d}| 
\end{equation} 
We keep MSE/MAE losses in evaluation to ensure comparability with baseline models. For the sake of fairness, We also evaluate the performance of certain baseline models trained with Arctan Weighted Loss in Section~\ref{Ablation Arctan}.
\section{Experiment}
\label{sec:experiments}

\subsection{Performance on Diverse Non-stationary Benchmarks}
Table~\ref{tab:results_main} provides a comprehensive evaluation of DMamba against nine competitive baselines across ETT, Electricity, Exchange, and Weather datasets. The baseline suite represents three dominant paradigms in LTSF: \textbf{Mamba-based}: S-Mamba\cite{smamba}, TimePro\cite{ma2025timepro}, \textbf{Transformer-based}: iTransformer\cite{liu2023itransformer}, PatchTST\cite{patchtst}, FEDformer\cite{fedformer}, TimesNet\cite{timesnet}, and \textbf{Linear/MLP-based}: RLinear\cite{rlinear}, DLinear\cite{dlinear}, TiDE\cite{tide}, XPatch\cite{xpatch}. 

The results demonstrate a significant breakthrough: DMamba consistently achieves SOTA performance across almost all horizons, particularly on the non-stationary ETT series. For instance, on ETTh2, DMamba outperforms \textit{S-Mamba} by 16.5\% with respect to average MSE. This validates our claim that entangled temporal patterns could overwhelm the Mamba module. Furthermore, DMamba consistently outperforms the recent baseline \textit{TimePro} in average metrics across most datasets, particularly on the ETTh2 and Weather benchmarks, where it achieves a significant margin. Even when compared to advanced Transformers like \textit{iTransformer} and \textit{PatchTST}, DMamba maintains a superior edge by leveraging Mamba's content-aware selective scan on purified seasonal components. Notably, DMamba also outperforms specialized decomposition models such as \textit{DLinear}, \textit{XPatch}, and \textit{FEDformer}. While these models utilize moving average or frequency-based decomposition, they often suffer from boundary padding bias or lose fine-grained temporal details. In contrast, DMamba’s EMA-based dual-stream architecture ensures a more robust decoupling of stable trends and high-frequency fluctuations, setting a new benchmark for long-term forecasting.

\subsection{Performance on PEMS Datasets}
To further assess spatial-temporal modeling capabilities, we evaluate DMamba on the PEMS (03, 04, 07, 08)\cite{pems1}\cite{pems2} benchmarks, which are characterized by high-dimensional variates and strong periodicity. As shown in Table~\ref{tab:results_pems}, DMamba yields consistent improvements even in scenarios where baselines like \textit{S-Mamba} and \textit{iTransformer} traditionally perform well. 

Compared to MLP-based architectures such as \textit{TiDE} and linear mappings like \textit{RLinear} and \textit{DLinear}—which often struggle with the complex, non-linear dependencies in traffic flow—DMamba exhibits superior stability. For example, on the PEMS08 dataset, DMamba reduces the average MSE from 0.148 (S-Mamba) and 0.150 (iTransformer) to an impressive 0.099, representing a 33.1\% improvement over its Mamba-based predecessor. Even against \textit{TimesNet} and \textit{PatchTST}, which excel at capturing local patterns, DMamba provides more accurate long-range forecasts. This success is attributed to the synergistic effect of the dual-stream network: while the trend stream handles stable shifts, the Mamba-based seasonal stream captures intricate inter-variate correlations free from the interference of low-frequency noise. These consistent gains across PEMS benchmarks solidify DMamba’s position as a superior general-purpose architecture for multivariate time series forecasting.

\subsection{Efficiency Analysis}
\begin{figure}[t]
    \centering
    \includegraphics[width=0.8\linewidth]{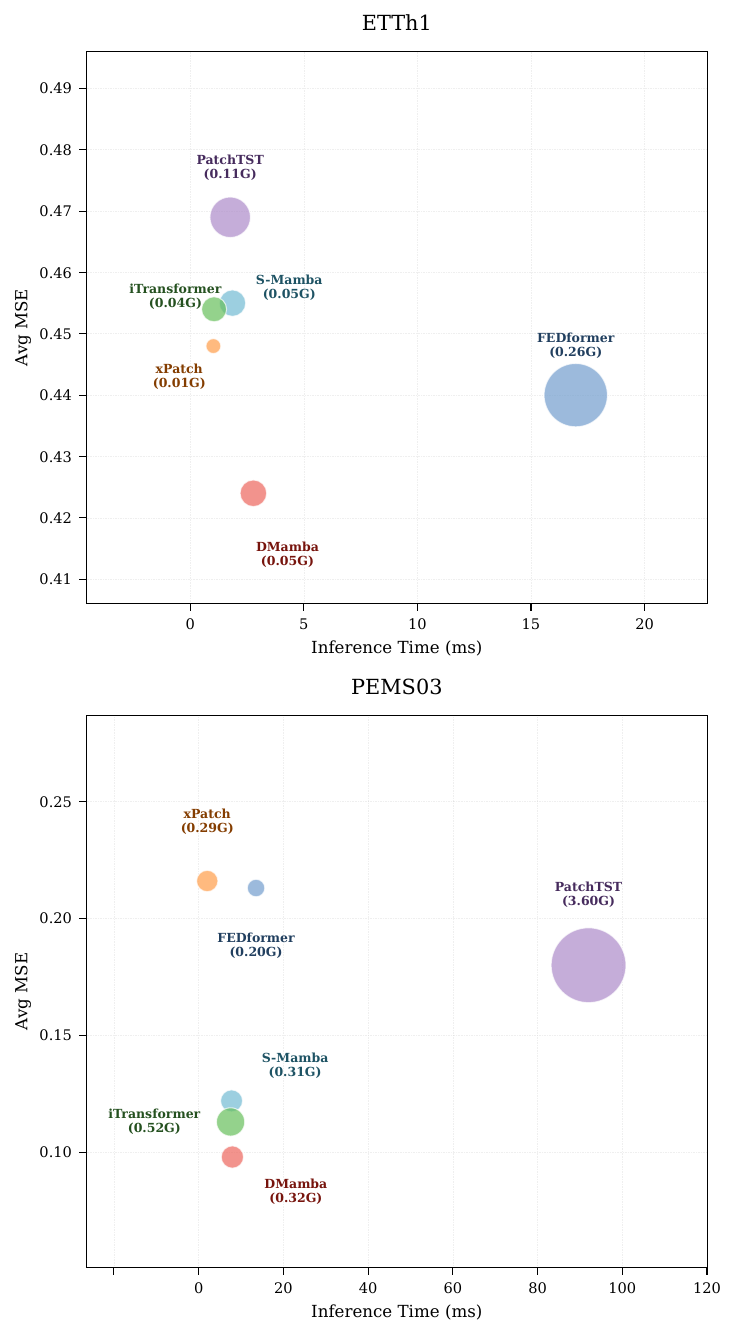}
    \caption{Efficiency trade-off on ETTh1 and PEMS03 with a look-back window of $L=96$. DMamba (bottom-left) achieves the best balance of low error (Avg MSE), fast inference, and low computational cost (FLOPs, bubble size).}
    \label{fig:efficiency_comparison}
\end{figure}

We evaluate the trade-off between predictive accuracy (Avg MSE), inference latency, and computational cost (FLOPs), as shown in Figure~\ref{fig:efficiency_comparison}. \textbf{DMamba} consistently occupies the optimal bottom-left quadrant, demonstrating a superior balance of high accuracy and efficiency. For instance, on PEMS03, DMamba is nearly \textbf{9x faster} than PatchTST (10ms vs. 90ms) and over \textbf{10x lighter} in FLOPs (0.32G vs. 3.60G), while achieving a lower MSE. This advantage stems from our hybrid design, which pairs the linear-time complexity of Mamba with a lightweight MLP, outperforming both high-latency Transformers and less accurate fast models.

\subsection{Ablation Study}
\subsubsection{Ablation Study on Branch Architectures across ETT
Datasets}
To evaluate the individual contributions of our hybrid design, we compare the proposed \textbf{DMamba} with four architectural variants: \textbf{All-MLP} (MLP for both branches), \textbf{Mamba-Trend} (MLP for Seasonal, Mamba for Trend), \textbf{All-Mamba} (Mamba for both branches), and \textbf{T-Mamba} (Temporal Mamba for Seasonal). The results summarized in Table~\ref{tab:ablation_final_updated} demonstrate that \textbf{DMamba} consistently achieves the lowest average MSE and MAE across all four ETT benchmarks. This validates our core strategy of utilizing high-capacity Mamba blocks for purified seasonal residuals while maintaining a stable MLP for trend projections to ensure overall robustness.

Analysis of individual datasets reveals that while \textbf{All-MLP} remains competitive on the relatively stable ETTh1 dataset—securing the best results for specific horizons such as $T=96$ and $T=720$—it generally lacks the representational power to model more complex temporal fluctuations. By introducing the Mamba block into the seasonal stream, \textbf{DMamba} effectively captures non-linear periodicities, leading to a substantial performance boost in the more non-stationary ETTh2 dataset. This confirms that the selective scan mechanism of Mamba is particularly potent when applied to high-frequency seasonal components.

The suboptimality of the \textbf{Mamba-Trend} configuration further reinforces our architectural choice. By assigning the Mamba block to the trend branch and an MLP to the seasonal branch, performance degrades significantly across almost all tested horizons. For instance, the average MSE for Mamba-Trend on ETTh2 is 0.355, which is notably inferior to \textbf{DMamba}'s 0.318. This clear performance gap demonstrates that the state-space paradigm is fundamentally better suited to process seasonal residuals than to model inter-channel correlations among trend components.

Furthermore, we evaluate \textbf{T-Mamba}, where the Mamba block scans along the temporal dimension rather than the variate dimension. \textbf{DMamba} surpasses \textbf{T-Mamba} across most metrics (e.g., improving Avg MSE from 0.365 to 0.318 on ETTh2), indicating that explicitly modeling inter-variate dependencies via channel-wise scanning is more critical than temporal scanning for capturing the dynamic interactions within seasonal fluctuations.

Finally, although the \textbf{All-Mamba} variant shows strong modeling capacity and achieves the best MSE on the ETTh2 horizon of $T=720$, its performance on longer horizons in ETTm1 and ETTm2 is less consistent compared to \textbf{DMamba}. This suggests that over-parameterizing the trend component with a high-capacity state-space model may lead to the capture of spurious correlations. \textbf{DMamba} maintains a superior balance by utilizing the MLP as a robust inductive bias for trend modeling, which prevents overfitting and enhances generalization across diverse non-stationary sequences. Consequently, the hybrid configuration of \textbf{DMamba} proves to be the most reliable architecture for robust long-term forecasting.

\subsubsection{Ablation Study on the Arctan Weighted Loss for DMamba} 
\label{Ablation Arctan}
To verify that the superior performance of \textbf{DMamba} stems from its architectural design rather than solely from the choice of loss function, we conduct a controlled experiment by equipping all models with the same \textbf{Arctan Weighted Loss}. We deliberately selected four of the most competitive recent baselines—S-Mamba, PatchTST, iTransformer, and xPatch—which represent the state-of-the-art in State Space Models, Transformers, and linear-based architectures. 

In Table~\ref{tab:loss_ablation_necessity}, models marked with an asterisk (*) indicate that their original uniform MSE objective has been replaced by this weighted loss. This setup ensures a fair comparison where the loss function is a controlled variable across these highly competitive backbones. 

The results demonstrate that while the Arctan Weighted Loss is a vital component of the \textbf{DMamba} framework, it does not compensate for the structural limitations of other architectures. Even when utilizing the same temporal prioritization, \textbf{DMamba*} consistently outperforms these strong baselines. This confirms that our architectural synergy—the combination of EMA-based decomposition and Mamba-based seasonal modeling—is the primary driver of robustness, with the specialized loss function serving as a complementary enhancement to this structural core.

\begin{table}[!h]
  \centering
  \caption{Performance comparison under a unified Arctan Weighted Loss across competitive baselines. Models marked with an asterisk (*) denote that their native uniform MSE loss has been substituted with the proposed weighted loss. \textbf{Bold} indicates the best performance, and \underline{underlined} indicates the second best.}
  \label{tab:loss_ablation_necessity}
  \renewcommand{\arraystretch}{1.1}
  \setlength{\tabcolsep}{5pt} 
  \resizebox{\columnwidth}{!}{
  \begin{tabular}{lc ccccc}
    \toprule
    \multicolumn{2}{c}{Settings} & \multicolumn{5}{c}{Models with Arctan Weighted Loss (*)} \\
    \cmidrule(lr){3-7}
    Dataset & Metric & \textbf{DMamba}* & S-Mamba* & PatchTST* & iTransformer* & xPatch* \\
    \midrule
    \multirow{2}{*}{ETTh1} 
     & MSE & \textbf{0.431} & 0.471 & \underline{0.446} & 0.448 & 0.448 \\
     & MAE & \textbf{0.426} & 0.451 & \underline{0.432} & 0.439 & 0.437 \\
    \midrule
    \multirow{2}{*}{ETTh2} 
     & MSE & \textbf{0.318} & 0.371 & 0.362 & 0.380 & \textbf{0.318} \\
     & MAE & \textbf{0.359} & 0.392 & 0.388 & 0.398 & \underline{0.363} \\
    \midrule
    \multirow{2}{*}{ETTm1} 
     & MSE & \textbf{0.376} & 0.403 & \underline{0.382} & 0.398 & 0.385 \\
     & MAE & \underline{0.385} & 0.397 & \textbf{0.380} & 0.393 & 0.388 \\
    \midrule
    \multirow{2}{*}{ETTm2} 
     & MSE & \textbf{0.266} & 0.282 & 0.281 & 0.283 & \underline{0.268} \\
     & MAE & \textbf{0.313} & 0.322 & 0.319 & 0.322 & \underline{0.315} \\
    \bottomrule
  \end{tabular}
  }
\end{table}

\subsubsection{Ablation Study on the Sensitivity Analysis of EMA Smoothing Factor} 
Our sensitivity analysis of the EMA smoothing factor $\alpha$ across ETT datasets shows its optimal value is data-dependent. ETTh1 benefits from a higher $\alpha$ (0.7) for agile adaptation, while $\alpha=0.3$ proved most robust for ETTh2, ETTm1, and ETTm2, effectively balancing trend stability and responsiveness. The EMA decomposition module demonstrates overall robustness within $\alpha \in [0.3, 0.7]$. Detailed results are in Table~\ref{tab:ema_sensitivity_v2} in the appendix.

\bibliographystyle{ACM-Reference-Format}
\bibliography{sample-base}

@String{Springer = "Springer-Verlag" }

@inproceedings{dlinear,
  title={Are transformers effective for time series forecasting?},
  author={Zeng, Ailing and Chen, Muxi and Zhang, Lei and Xu, Qiang},
  booktitle={Proceedings of the AAAI conference on artificial intelligence},
  volume={37},
  number={9},
  pages={11121--11128},
  year={2023}
}

@article{tide,
  title={Long-term forecasting with tide: Time-series dense encoder},
  author={Das, Abhimanyu and Kong, Weihao and Leach, Andrew and Mathur, Shaan and Sen, Rajat and Yu, Rose},
  journal={arXiv preprint arXiv:2304.08424},
  year={2023}
}

@article{rlinear,
  title={Revisiting long-term time series forecasting: An investigation on linear mapping},
  author={Li, Zhe and Qi, Shiyi and Li, Yiduo and Xu, Zenglin},
  journal={arXiv preprint arXiv:2305.10721},
  year={2023}
}

@article{autoformer,
  title={Autoformer: Decomposition transformers with auto-correlation for long-term series forecasting},
  author={Wu, Haixu and Xu, Jiehui and Wang, Jianmin and Long, Mingsheng},
  journal={Advances in neural information processing systems},
  volume={34},
  pages={22419--22430},
  year={2021}
}

@inproceedings{fedformer,
  title={Fedformer: Frequency enhanced decomposed transformer for long-term series forecasting},
  author={Zhou, Tian and Ma, Ziqing and Wen, Qingsong and Wang, Xue and Sun, Liang and Jin, Rong},
  booktitle={International conference on machine learning},
  pages={27268--27286},
  year={2022},
  organization={PMLR}
}

@article{patchtst,
  title={A Time Series is Worth 64Words: Long-term Forecasting with Transformers},
  author={Nie, Y},
  journal={arXiv preprint arXiv:2211.14730},
  year={2022}
}

@inproceedings{ett,
  title={Informer: Beyond efficient transformer for long sequence time-series forecasting},
  author={Zhou, Haoyi and Zhang, Shanghang and Peng, Jieqi and Zhang, Shuai and Li, Jianxin and Xiong, Hui and Zhang, Wancai},
  booktitle={Proceedings of the AAAI conference on artificial intelligence},
  volume={35},
  number={12},
  pages={11106--11115},
  year={2021}
}

@inproceedings{mamba,
  title={Mamba: Linear-time sequence modeling with selective state spaces},
  author={Gu, Albert and Dao, Tri},
  booktitle={First conference on language modeling},
  year={2024}
}

@article{s4,
  title={Efficiently modeling long sequences with structured state spaces},
  author={Gu, Albert and Goel, Karan and R{\'e}, Christopher},
  journal={arXiv preprint arXiv:2111.00396},
  year={2021}
}

@article{smamba,
  title={Is mamba effective for time series forecasting?},
  author={Wang, Zihan and Kong, Fanheng and Feng, Shi and Wang, Ming and Yang, Xiaocui and Zhao, Han and Wang, Daling and Zhang, Yifei},
  journal={Neurocomputing},
  volume={619},
  pages={129178},
  year={2025},
  publisher={Elsevier}
}

@inproceedings{revin,
  title={Reversible instance normalization for accurate time-series forecasting against distribution shift},
  author={Kim, Taesung and Kim, Jinhee and Tae, Yunwon and Park, Cheonbok and Choi, Jang-Ho and Choo, Jaegul},
  booktitle={International conference on learning representations},
  year={2021}
}

@inproceedings{xpatch,
  title={xPatch: Dual-Stream Time Series Forecasting with Exponential Seasonal-Trend Decomposition},
  author={Stitsyuk, Artyom and Choi, Jaesik},
  booktitle={Proceedings of the AAAI Conference on Artificial Intelligence},
  volume={39},
  number={19},
  pages={20601--20609},
  year={2025}
}

@article{ema,
  title={An exponential moving-average sequence and point process (EMA1)},
  author={Lawrance, AJ and Lewis, PAW},
  journal={Journal of Applied Probability},
  volume={14},
  number={1},
  pages={98--113},
  year={1977},
  publisher={Cambridge University Press}
}

@article{transformerforecasting,
  title={A systematic review for transformer-based long-term series forecasting},
  author={Su, Liyilei and Zuo, Xumin and Li, Rui and Wang, Xin and Zhao, Heng and Huang, Bingding},
  journal={Artificial Intelligence Review},
  volume={58},
  number={3},
  pages={80},
  year={2025},
  publisher={Springer}
}

@article{timesnet,
  title={Timesnet: Temporal 2d-variation modeling for general time series analysis},
  author={Wu, Haixu and Hu, Tengge and Liu, Yong and Zhou, Hang and Wang, Jianmin and Long, Mingsheng},
  journal={arXiv preprint arXiv:2210.02186},
  year={2022}
}

@article{etsformer,
  title={Etsformer: Exponential smoothing transformers for time-series forecasting},
  author={Woo, Gerald and Liu, Chenghao and Sahoo, Doyen and Kumar, Akshat and Hoi, Steven},
  journal={arXiv preprint arXiv:2202.01381},
  year={2022}
}

@article{card,
  title={Card: Channel aligned robust blend transformer for time series forecasting},
  author={Xue, Wang and Zhou, Tian and Wen, Qingsong and Gao, Jinyang and Ding, Bolin and Jin, Rong},
  journal={arXiv preprint arXiv:2305.12095},
  year={2023}
}

@article{patro2024simba,
  title={Simba: Simplified mamba-based architecture for vision and multivariate time series},
  author={Patro, Badri N and Agneeswaran, Vijay S},
  journal={arXiv preprint arXiv:2403.15360},
  year={2024}
}

@article{gu2020hippo,
  title={Hippo: Recurrent memory with optimal polynomial projections},
  author={Gu, Albert and Dao, Tri and Ermon, Stefano and Rudra, Atri and R{\'e}, Christopher},
  journal={Advances in neural information processing systems},
  volume={33},
  pages={1474--1487},
  year={2020}
}

@article{Discretization,
  title={Combining recurrent, convolutional, and continuous-time models with linear state space layers},
  author={Gu, Albert and Johnson, Isys and Goel, Karan and Saab, Khaled and Dao, Tri and Rudra, Atri and R{\'e}, Christopher},
  journal={Advances in neural information processing systems},
  volume={34},
  pages={572--585},
  year={2021}
}

@article{bimamba,
  title={Bi-mamba+: Bidirectional mamba for time series forecasting},
  author={Liang, Aobo and Jiang, Xingguo and Sun, Yan and Shi, Xiaohou and Li, Ke},
  journal={arXiv preprint arXiv:2404.15772},
  year={2024}
}

@article{liu2023itransformer,
  title={itransformer: Inverted transformers are effective for time series forecasting},
  author={Liu, Yong and Hu, Tengge and Zhang, Haoran and Wu, Haixu and Wang, Shiyu and Ma, Lintao and Long, Mingsheng},
  journal={arXiv preprint arXiv:2310.06625},
  year={2023}
}

@article{pems1,
  title={Spatio-temporal graph convolutional networks: A deep learning framework for traffic forecasting},
  author={Yu, Bing and Yin, Haoteng and Zhu, Zhanxing},
  journal={arXiv preprint arXiv:1709.04875},
  year={2017}
}

@inproceedings{pems2,
  title={Attention based spatial-temporal graph convolutional networks for traffic flow forecasting},
  author={Guo, Shengnan and Lin, Youfang and Feng, Ning and Song, Chao and Wan, Huaiyu},
  booktitle={Proceedings of the AAAI conference on artificial intelligence},
  volume={33},
  number={01},
  pages={922--929},
  year={2019}
}

@inproceedings{ahamed2024timemachine,
  title={Timemachine: A time series is worth 4 mambas for long-term forecasting},
  author={Ahamed, Md Atik and Cheng, Qiang},
  booktitle={ECAI 2024: 27th European Conference on Artificial Intelligence, 19-24 October 2024, Santiago de Compostela, Spain-Including 13th Conference on Prestigious Applications of Intelligent Systems. European Conference on Artificial Intelli},
  volume={392},
  pages={1688},
  year={2024}
}

@article{inoue2021short,
  title={Short-term traffic speed prediction based on fundamental and cointegration relationship of speed--density in non-congested and congested states},
  author={Inoue, Ryo and Miyashita, Akihisa},
  journal={IEEE Open Journal of Intelligent Transportation Systems},
  volume={2},
  pages={470--481},
  year={2021},
  publisher={IEEE}
}

@article{engle1987co,
  title={Co-integration and error correction: representation, estimation, and testing},
  author={Engle, Robert F and Granger, Clive WJ},
  journal={Econometrica: journal of the Econometric Society},
  pages={251--276},
  year={1987},
  publisher={JSTOR}
}

@article{johansen1990maximum,
  title={Maximum likelihood estimation and inference on cointegration—with appucations to the demand for money},
  author={Johansen, Soren and Juselius, Katarina},
  journal={Oxford Bulletin of Economics and statistics},
  volume={52},
  number={2},
  pages={169--210},
  year={1990}
}

@article{gatev2006pairs,
  title={Pairs trading: Performance of a relative-value arbitrage rule},
  author={Gatev, Evan and Goetzmann, William N and Rouwenhorst, K Geert},
  journal={The review of financial studies},
  volume={19},
  number={3},
  pages={797--827},
  year={2006},
  publisher={Oxford University Press}
}

@article{dwyer1992cointegration,
  title={Cointegration and market efficiency},
  author={Dwyer Jr, Gerald P and Wallace, Myles S},
  journal={Journal of International Money and Finance},
  volume={11},
  number={4},
  pages={318--327},
  year={1992},
  publisher={Elsevier}
}

@article{siliverstovs2005international,
  title={International market integration for natural gas? A cointegration analysis of prices in Europe, North America and Japan},
  author={Siliverstovs, Boriss and L'H{\'e}garet, Guillaume and Neumann, Anne and Von Hirschhausen, Christian},
  journal={Energy Economics},
  volume={27},
  number={4},
  pages={603--615},
  year={2005},
  publisher={Elsevier}
}

@article{kaufmann2002cointegration,
  title={Cointegration analysis of hemispheric temperature relations},
  author={Kaufmann, Robert K and Stern, David I},
  journal={Journal of Geophysical Research: Atmospheres},
  volume={107},
  number={D2},
  pages={ACL--8},
  year={2002},
  publisher={Wiley Online Library}
}

@article{ma2025timepro,
  title={TimePro: Efficient Multivariate Long-term Time Series Forecasting with Variable-and Time-Aware Hyper-state},
  author={Ma, Xiaowen and Ni, Zhenliang and Xiao, Shuai and Chen, Xinghao},
  journal={arXiv preprint arXiv:2505.20774},
  year={2025}
}

\appendix

\section{Ablation Study on the Sensitivity Analysis of EMA Smoothing Factor}
To investigate the impact of the EMA smoothing factor $\alpha$ on forecasting performance, we conduct a sensitivity analysis across all ETT datasets. The smoothing factor $\alpha$ is a critical hyperparameter that dictates the trade-off between trend stability and responsiveness. As shown in Table~\ref{tab:ema_sensitivity_v2}, the optimal value for $\alpha$ varies slightly depending on the data characteristics.

For the ETTh1 dataset, a larger smoothing factor ($\alpha=0.7$) achieves the best performance, indicating that this dataset benefits from a more agile trend adaptation that can quickly respond to shifts. In contrast, for ETTh2, ETTm1, and ETTm2, the baseline value of $\alpha=0.3$ remains the most robust choice, consistently yielding the lowest MSE and MAE. This suggests that for many non-stationary scenarios, a moderate $\alpha$ effectively prevents the trend component from over-fitting to high-frequency noise while still capturing long-term shifts. The relatively stable performance across the range of $\alpha \in [0.3, 0.7]$ demonstrates the inherent robustness of our EMA-based decomposition module.

\begin{table}[htbp]
  \caption{Sensitivity analysis of the EMA smoothing factor $\alpha$ across ETT datasets ($L=96, T=96$). We evaluate the performance impact of different $\alpha$ values. \textbf{Bold} indicates the best performance, and \underline{underlined} indicates the second best.}
  \label{tab:ema_sensitivity_v2}
  \centering
  \renewcommand{\arraystretch}{1.0}
  \setlength{\tabcolsep}{4pt} 
  \resizebox{\columnwidth}{!}{
  \begin{tabular}{lc ccccc c}
    \toprule
    \multicolumn{2}{c}{Settings} & \multicolumn{5}{c}{Smoothing Factor $\alpha$} & \multirow{2}{*}{Best $\alpha$} \\
    \cmidrule(lr){3-7}
    Dataset & Metric & 0.1 & 0.3 & 0.5 & 0.7 & 0.9 & \\
    \midrule
    \multirow{2}{*}{ETTh1} 
     & MSE & 0.380 & 0.377 & \underline{0.375} & \textbf{0.374} & 0.377 & 0.7 \\
     & MAE & 0.394 & 0.390 & \underline{0.387} & \textbf{0.386} & \textbf{0.386} & \\
    \midrule
    \multirow{2}{*}{ETTh2} 
     & MSE & \underline{0.287} & \textbf{0.235} & 0.290 & 0.290 & 0.288 & 0.3 \\
     & MAE & \underline{0.333} & \textbf{0.301} & 0.336 & 0.336 & \underline{0.333} & \\
    \midrule
    \multirow{2}{*}{ETTm1} 
     & MSE & 0.316 & \textbf{0.307} & 0.312 & \underline{0.309} & 0.313 & 0.3 \\
     & MAE & 0.343 & 0.345 & \underline{0.341} & \textbf{0.338} & \textbf{0.338} & \\
    \midrule
    \multirow{2}{*}{ETTm2} 
     & MSE & 0.173 & \textbf{0.164} & 0.172 & \underline{0.171} & 0.172 & 0.3 \\
     & MAE & 0.253 & \textbf{0.247} & 0.251 & \underline{0.250} & 0.251 & \\
    \bottomrule
  \end{tabular}
  }
\end{table}

\end{document}